%% file: main.tex
\documentclass[10pt,twocolumn,letterpaper]{article}
\usepackage{iccv}
\usepackage{times}
\usepackage{epsfig}
\usepackage{graphicx}
\usepackage{amsmath}
\usepackage{amssymb}
\usepackage{float}
\usepackage{color}
\usepackage[clock]{ifsym}
\usepackage{booktabs}

\usepackage{microtype}

\usepackage[breaklinks=true,bookmarks=false]{hyperref}

\iccvfinalcopy %

\def\iccvPaperID{1868} %

\begin{document}

\title{Predicting the Category and Attributes of Visual Search Targets\\Using Deep Gaze Pooling}
\author{
Hosnieh Sattar$^{1,2}$
\and
Andreas Bulling$^1$\\
\and
Mario Fritz$^2$
\and
$^1$Perceptual User Interfaces Group, $^2$Scalable Learning and Perception Group\\
Max Planck Institute for Informatics, Saarland Informatics Campus, Saarbr\"ucken, Germany\\
{\tt\small \{sattar,mfritz,bulling\}@mpi-inf.mpg.de}
}
\maketitle

\input{abstract.tex}

\input{Introduction.tex}

\input{Relatedworks.tex}

\input{Datacollection.tex}

\input{Methods.tex}

\input{ExperimentsandResults.tex}

\input{Conclusions.tex}
{\small
\bibliographystyle{ieee}
\bibliography{egbib}
}
\input{sup.tex}

\end{document}

%% file: abstract.tex
\begin{abstract}

Predicting the target of visual search from eye fixation (gaze) data is a challenging problem with many applications in human-computer interaction. 
In contrast to previous work that has focused on individual instances as a search target, we propose the first approach to predict categories and attributes of search targets based on gaze data.
However, state of the art models for categorical recognition, in general, require large amounts of training data, which is prohibitive for gaze data.
To address this challenge, we propose a novel {\it Gaze Pooling Layer} that integrates gaze information into CNN-based architectures as an attention mechanism -- incorporating both spatial and temporal aspects of human gaze behavior.
We show that our approach is effective even when the {\it gaze pooling layer} is added to an already trained CNN, thus eliminating the need for expensive joint data collection of visual and gaze data.
We propose an experimental setup and data set and demonstrate the effectiveness of our method for search target prediction based on gaze behavior.
We further study how to integrate temporal and spatial gaze information most effectively, and indicate directions for future research in the gaze-based prediction of mental states.

\end{abstract}

%% file: Introduction.tex
\section{Introduction}

\begin{figure}[t]
\centering
\includegraphics[width=\columnwidth]{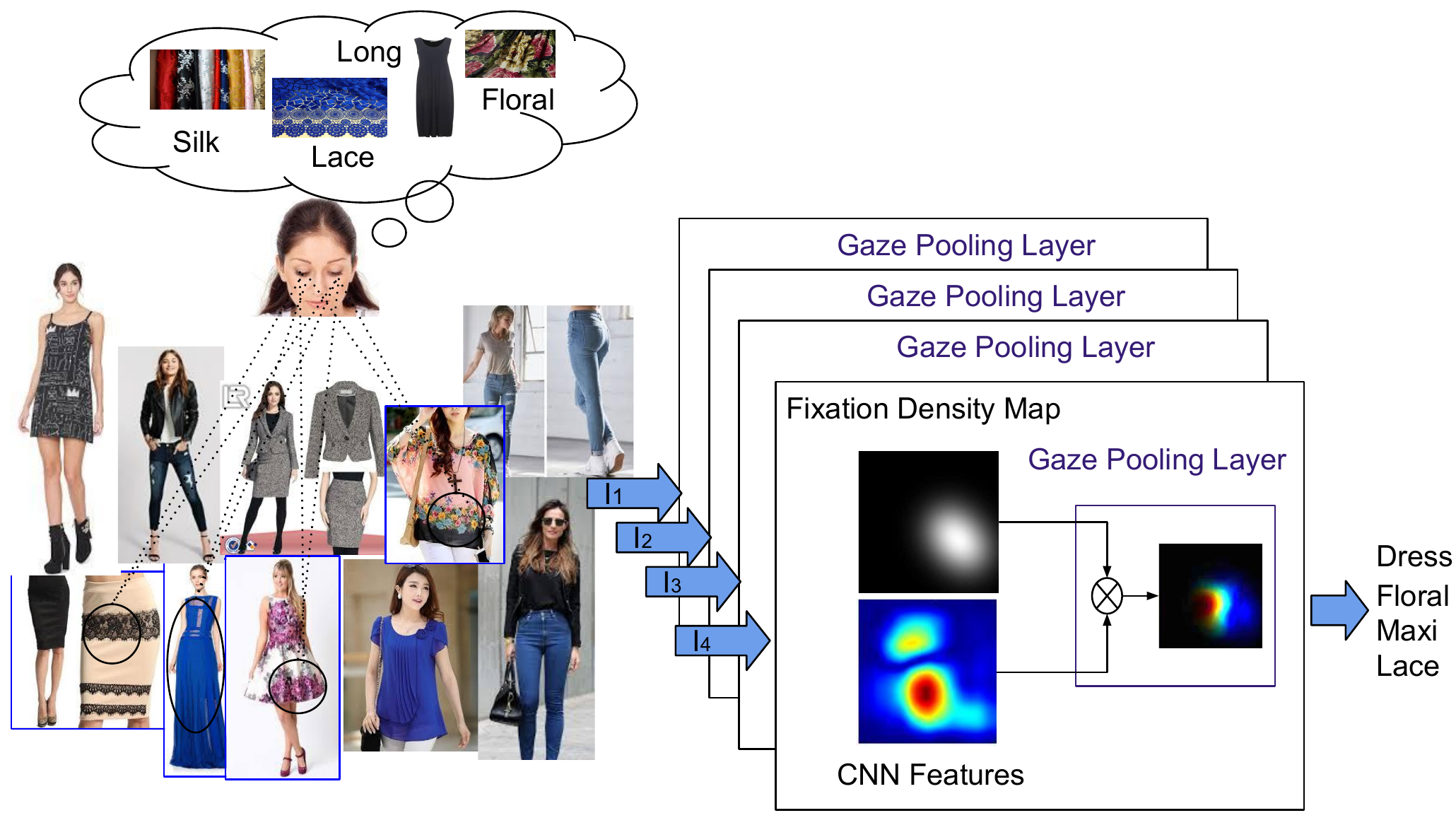}
\caption{We propose a method to predict the target of visual search in terms of categories and attributes from users' gaze. We propose a {\it Gaze Pooling Layer} that leverages gaze data as an attention mechanism in a trained CNN architecture.} 
\label{fig:teaser}
\end{figure}

As eye tracking technology is beginning to mature, there is an increasing interest in exploring the type of information that can be extracted from human gaze data.
Within the wider scope of eye-based activity recognition~\cite{bulling11_pami,steil2015discovery}, search target prediction \cite{borji2014eyes, sattar15_cvpr, zelinsky2013eye} has recently received particular attention as it aims to recognize users' search intents without the need for them to verbally communicate these intends. 
Previous work on search target prediction from gaze data (e.g. \cite{borji2014eyes, sattar15_cvpr}) is limited to specific target instances that users searched for, e.g.\ a particular object.
This excludes searches for broader classes of objects that share the same semantic category or certain object attributes.
Such searches commonly occur if the user does not have a concrete target instance in mind but is only looking for an object from a certain category or with certain characteristic attributes.

To address these limitations, we broaden the scope of search target prediction to categorical classes, such as object categories or attributes.
One key difficulty towards achieving this goal is acquiring sufficient training data.
We have to recall that object categorization only in the past decade has seen a breakthrough in performance by combining deep learning techniques with large training corpora.
Collecting such large corpora is prohibitive for human gaze data, which poses a severe challenge to achieve our goal.

Therefore, we propose an approach for predicting categories and attributes of search targets that utilize readily trained CNN architectures and combines them with gaze data in a novel {\it Gaze Pooling Layer} (see Figure~\ref{fig:teaser}).
The gaze information is used as an attention mechanism that acts selectively on the visual features to predict users' search target.
These design choices make our approach compatible and practical with current deep learning architectures.

Through extensive experiments, we show that our method achieves accurate search target prediction for 10 category and 10 attribute tasks on a new gaze data set that is based on the DeepFashion data set~\cite{liuLQWTcvpr16DeepFashion}.
Furthermore, we evaluate different parameter settings and design choices of our approach, visualize internal representations and perform a robustness study w.r.t. noise in the eye tracking data.
All code and data will be made publicly available upon acceptance.

%% file: Relatedworks.tex
\section{Related Work}

Predicting the target of visual search is a task studied both in computer vision
\cite{Kovashka2015,borji2014eyes,4657362,sattar15_cvpr,qian2016,zelinsky2013eye} and human
perception \cite{doi:10.1167/11.5.14,Chen20064118,Levin2001,Neider20062217}.
Existing approaches vary in the granularity of the predictions, either focusing on predicting specific
object instances \cite{borji2014eyes, sattar15_cvpr} or operating at the coarser level and
predicting target categories~\cite{4657362,zelinsky2013eye}.
The type of user feedback varies as well. While \cite{borji2014eyes,sattar15_cvpr,zelinsky2013eye} solely use implicit
information obtained from human gaze, \cite{Kovashka2015,4657362,qian2016}
require the user to provide explicit relevance feedback.
In the following, we summarize previous works on gaze-supported computer vision, user feedback for image search and retrieval, as well as methods for search target prediction.

\medskip

\vspace{0.1em}
\noindent\textbf{Gaze-Supported Computer Vision.} Our approach is related to an increasing
body of computer vision literature that employs gaze as a means to provide supervision or indicate
salient regions in the image in a variety of recognition tasks.  Visual fixations have been used in
\cite{li2014secrets,xu2014predicting} to indicate object locations in the context of saliency
predictions, and in ~\cite{karthikeyan2013,papadopoulos2013gaze,Shcherbatyi15_arxiv} as a form of weak supervision for the training of object detectors. Gaze information has been used to analyze pose estimation tasks in
~\cite{mps13iccv,subramanian2011can} as well as for action detection~\cite{mathe2014multiple}.
Gaze data has also been
employed for active segmentation ~\cite{mishra2009active}, localizing important objects in
egocentric videos~\cite{damen2014you,toyama2012gaze}, image captioning and scene
understanding \cite{SuganoB16}, as well as zero-shot image classification~\cite{karessli2017_cvpr}.
 
\medskip
\vspace{0.1em}\noindent\textbf{User Feedback for Image Search and Retrieval.} To close the semantic gap
between a user's envisioned search target and the images retrieved by search engines, Ferecatu and
Geman~\cite{4657362} proposed a framework to discover the semantic category of user's mental image
in unstructured data via explicit user input.
Kovashka et al.~\cite{Kovashka2015} introduced a novel
explicit feedback method to assess the mental models of users. Most recently Yu et
al.~\cite{qian2016} proposed to use free-hand human sketches as queries to perform instance-level
retrieval of images. They considered these sketches to be manifestations of users' mental model of
the target. The common theme in these approaches is that they require explicit user input as part of
their search refinement loop. Mouse clicks were used as input in \cite{4657362}. \cite{Kovashka2015}
used a set of attributes and required users to operate on a large attribute vocabulary to describe
their mental images.  In \cite{qian2016} the feedback was provided by sketching the target to
convey concepts such as texture, color, material, and style, which is a non-trivial step for most
users. 
In contrast, in our work, we do not rely on a feedback loop as in \cite{Kovashka2015} or
explicit user input or some form of initial description of a target as in
\cite{Kovashka2015,4657362,qian2016}. We instead use fixation information that can
be acquired implicitly during the search task itself, and demonstrate that such information allows
us to predict categories as well as attributes of search targets in a single search session.

\medskip
\vspace{0.1em}\noindent\textbf{Visual Search Target Prediction.} Human gaze
behavior reflects cognitive processes of the mind, such as
intentions~\cite{Brigham,kleinke:1986,Land1231}, and is influenced by the user's task
\cite{yarbus1967eye}. In the context of visual search, previous work typically focused on predicting
targets corresponding to specific object instances
\cite{borji2014eyes,sattar15_cvpr,zelinsky2013eye}.  For example, users were required to search for
specific book covers \cite{sattar15_cvpr} or specific binary patterns \cite{borji2014eyes} among
other distracting objects. In contrast, in this work, we aim to infer the general properties of a
search target represented by the object's category and attributes.
In this scenario, the search task is guided by the mental model that the user has the object class
rather than a specific instance of an object \cite{4657362,Wilson2008}. This presents additional
challenges as mental models might differ substantially among subjects. Furthermore,
\cite{borji2014eyes,sattar15_cvpr,zelinsky2013eye} required gaze data for training, whereas our
approach can be pre-trained on visual data alone, and then combined with gaze data at test time.

%% file: Datacollection.tex
\section{Data Collection}
\label{sec:data}

No existing data set provides image and gaze data that is suitable for our search target prediction task.
We, therefore, collected our own gaze data set based on the DeepFashion data set \cite{liuLQWTcvpr16DeepFashion}.
DeepFashion is a clothes data set consisting of 289,222 images annotated with 46 different categories and 1,000 attributes. 
We used the top 10 categories and attributes in our data collection.
The training set of DeepFashion was used to train our CNN image model for clothes category and attribute prediction;
the validation set was used to train participants for each category and attribute (see below). 
Finally, the test set was used to build up image collages for which we recorded human gaze data of participants while searching for specific categories and attributes.
In the following, we describe our data collection in more detail. 

\subsection{Participants and Apparatus}

We collected data from 14 participants (six females), aged between 18 and 30 years and with different nationalities.
All of them had normal or corrected-to-normal vision.
For gaze data collection we used a stationary Tobii TX300 eye tracker that provides binocular gaze data at a sampling frequency of 300Hz.
We calibrated the eye tracker using a standard 9-point calibration, followed by a validation of eye tracker accuracy.
For gaze data processing we used the Tobii software with the parameters for fixation detection left at their defaults (fixation duration: 60ms, the maximum time between fixations: 75ms).
Image collages were shown on a 30-inch screen with a resolution of 2560x1600 pixels.

\subsection{Procedure}

\begin{figure}[t]
\centering
\includegraphics[width=\columnwidth]{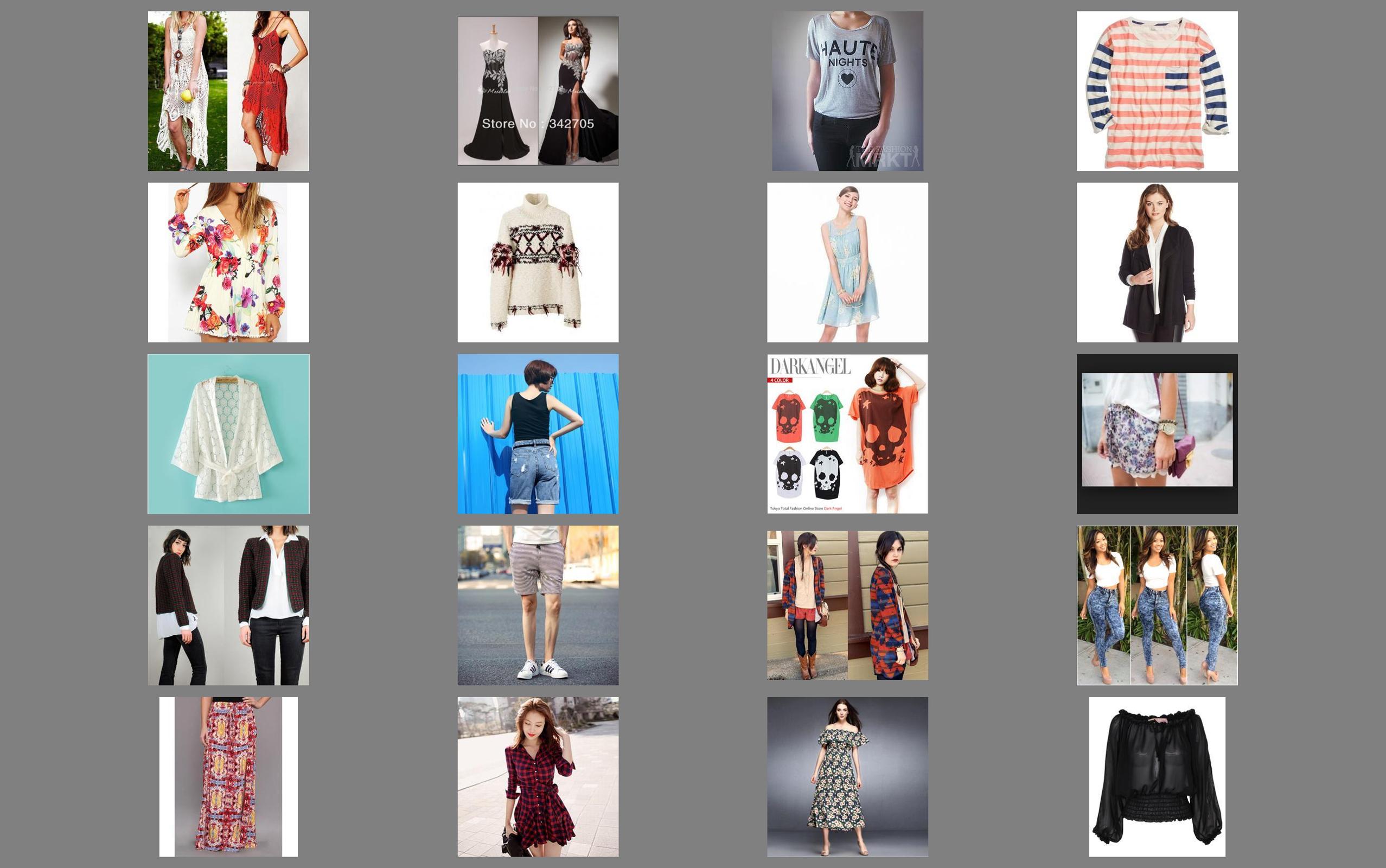}\\[0.1cm]
\includegraphics[width=\columnwidth]{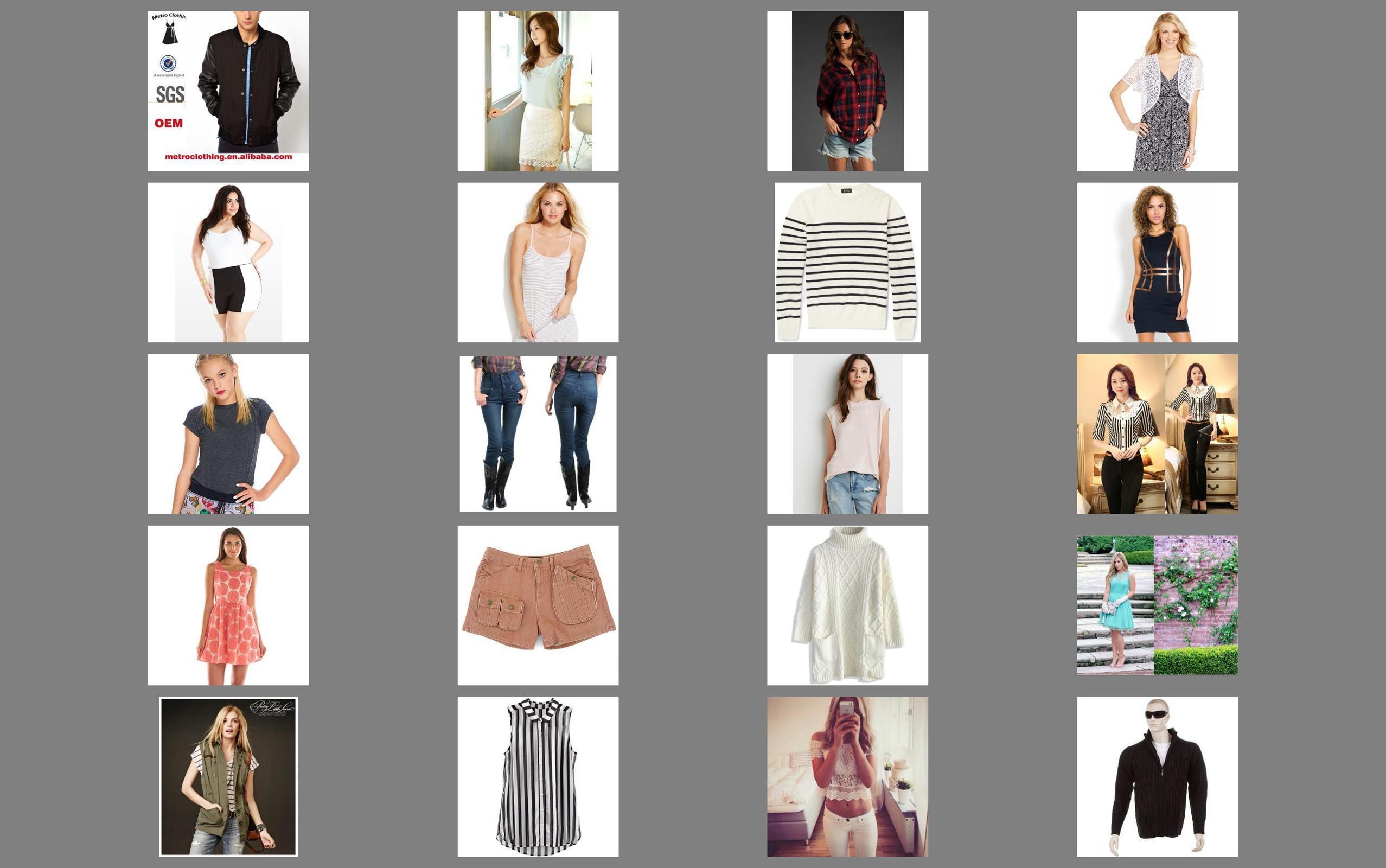}\\[0.1cm]
\caption{Sample image collages used for data collection: Attributes (top), Categories (bottom). Participants were asked to find different clothing attributes and categories within these collages.}
\label{fig:collages}
\end{figure}

We first trained participants by showing them exemplar images of all categories and attributes in a game like session to familiarize themselves with the categories and attributes.
We did not collect any gaze data at this stage.
For each category and attribute, we then generated 10 image collages, each containing 20 images.
Each target category or attribute appeared twice in each collage at a random location (see \autoref{fig:collages} for an example).
Participants were then asked to search for ten different categories and attributes on these image collages (see \autoref{fig:collages}) while their gaze was being tracked. 
We stress again that we did not show participants a specific target instance of a category or attribute that they should search for.
Instead, we only instructed them to find a matching image from a certain category, i.e\ ``dress'', or with a certain attribute, i.e\ ``floral''.
Consequently, search session guided by the mental image of participants from the specific category or attributes.
Participants had a maximum of 10 seconds to find the asked target category or attribute in the collage that was shown full-screen.
As soon as participants found a matching target, they were asked to press a key.
Afterward, they were asked whether they had found a matching target and how difficult the search had been.
This procedure was repeated ten times for ten different categories or attributes, resulting in a total of 100 search tasks.

%% file: Methods.tex
\begin{figure*}
\begin{center}
\includegraphics[width=1\linewidth]{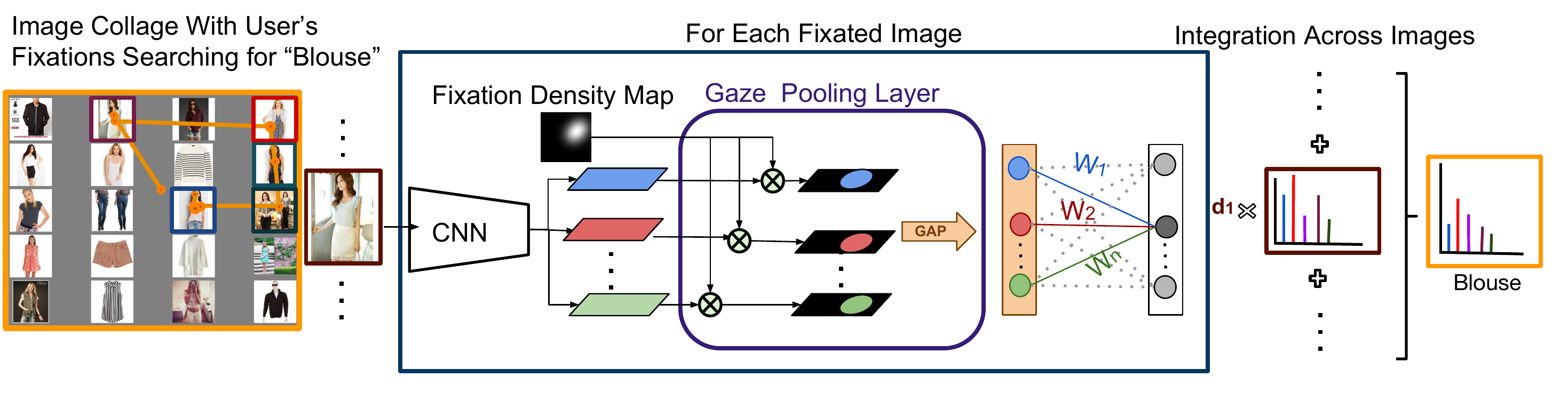}
\end{center}
   \caption{Overview of our approach. Given a search task (e.g. ``Find a blouse"), participants fixate on multiple images in an image collage. Each fixated image is encoded into multiple spatial features using a pre-trained CNN. The proposed Gaze Pooling Layer combines visual features and fixation density maps in a feature-weighting scheme. The output is a prediction of the category or attributes of the search target. To obtain one final prediction over image collages, we integrate the class posteriors across all fixated images using average pooling. 
 } 

\label{fig:short}
\end{figure*}
\begin{figure*}
\begin{center}
\includegraphics[width=\linewidth]{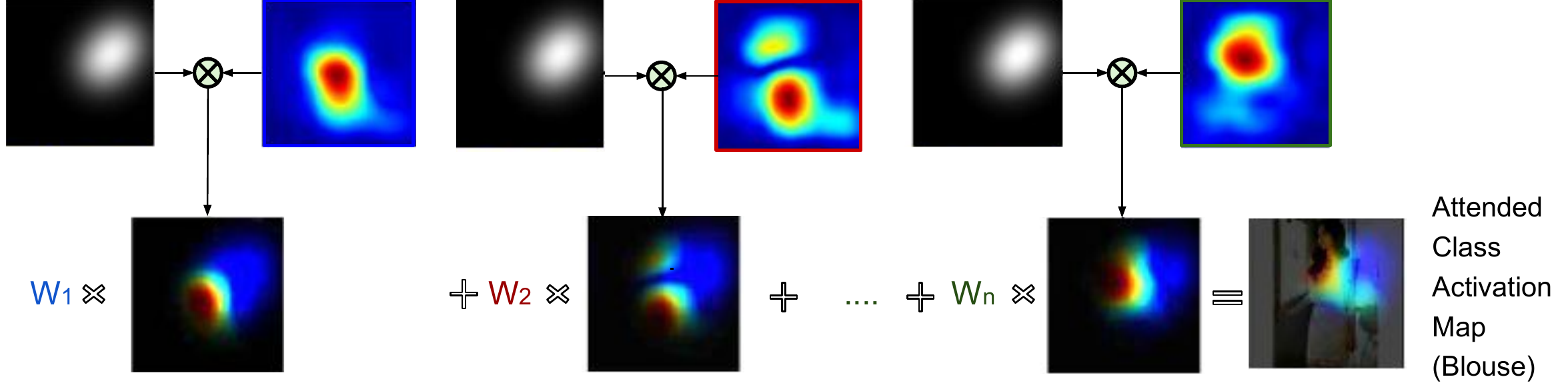}
\end{center}
\caption{The proposed Gaze Pooling Layer combines fixation density maps with CNN feature maps via a spatial re-weighting (top row). Attended class activation maps are shown in the bottom row, which the predicted class scores are mapped back to the previous convolutional layer. The attended class activation maps highlight the class-specific discriminative image regions.}
\label{fig:short2}
\end{figure*}
\section{Prediction of Search Targets Using Gaze}
In this work, we are interested in predicting the category and attributes of search targets from gaze data. %
We address this task by introducing the Gaze Pooling Layer (GPL) that combines CNN architectures with gaze data in a weighting mechanism. %
\autoref{fig:short} gives an overview of our approach.
In the following, we describe the four major components of our method in detail: The image encoder, human gaze encoding, the Gaze Pooling Layer, and search target prediction.
Finally, we also discuss different integration schemes across multiple images that allow us to utilize gaze information obtained from collages.
As a mean of inspecting the internal representation of our Gaze Pooling Layer, we propose Attended Class Activation Maps (ACAM).
\subsection{Image Encoder}
We build on the recent success of deep learning and use a convolutional neural network (CNN) to encode image information \cite{simonyan2013deep,NIPS2012_4824}. 
Given a raw image $I$, a CNN is used to extract image feature map $F(I)$.
\begin{align}
F(I)=\textbf{CNN}(I)
\end{align}
The end-to-end training properties of these networks allow us to obtain domain-specific features.
In our case, the network will be trained with data and labels relevant to the fashion domain.
As we are interested in combining spatial gaze features with the image features, we use features $F(I)$ of the last convolutional layer that still has a spatial resolution. 
This results in a task-dependent representation with spatial resolution.
In addition, to gain a higher spatial resolution we used the same architecture as describe in~\cite{zhou2016cvpr}.
We use their VGGnet-based model where layers after \texttt{conv5-3} are removed to gain a resolution of $14 \times 14$.
\subsection{Human Gaze Encoding}
\label{attention}
Given a target category or attributes, participant $P \in \mathbb{P}$ look at image $I$ and performs fixations $\textit{G(I,P)}={(x_i,y_i),i=1,...,N}$ in screen coordinates.
We aggregate these fixations into fixation density maps $FDM(G)$ that capture the spatial density of fixations over the full image.
Therefore, we represent the fixation density map $FDM(g)$ for a single fixation $g \in G(I,P)$  by a Gaussian:
\begin{align}
FDM(g) = \mathcal{N}(g,\sigma_{\text{fix}}),
\end{align}
centered at the coordinates of the fixation, with a fixed standard deviation $\sigma_{\text{fix}}$ -- the only parameter of our representation.
The fixation density map for all fixations $FDM(G)$ is obtained by coordinate-wise summation:  
\begin{align}
FDM(G) = \sum_{g \in G}{FDM(g)}
\end{align}
This corresponds to an average pooling integration.
We also propose a max pooling version as follows:
\begin{align}
FDM(G) = \max_{g \in G}{FDM(g)}
\end{align}
\subsection{Gaze Pooling Layer}
\label{gazepooling}
We combine the visual features $F(I)$ with fixation density map $FDM(G)$ in a Gaze Pooling Layer. 
The integration is performed by element-wise multiplication between both to obtain a gaze-weighted feature map (GWFM)
\begin{align}
\textit{GWFM}(I,G) = F(I) \otimes FDM(G).
\end{align}
In spirit of \cite{zhou2016cvpr}, we then perform Global Average Pooling (GAP) on each feature channel separately in order to yield a vector-valued feature representation.
\begin{align}
\textit{GAP}_\text{GWFM}(I,G)= \sum_{x,y} \textit{GWFM}(I,G)
\end{align}
We finish our pipeline by classification with a fully connected layer and a soft-max layer. 
\begin{align}
p(C | I,G) = \textit{softmax}(W \; \textit{GAP}_\text{GWFM}(I,G) +b),
\end{align}
where $W$ are the learned weights and $b$ is the bias and $C$ are the considered classes. 
The classes represent either categories or attributes depending on the experiment and we decide for the class with the highest class posterior (see \autoref{fig:short}).

\subsection{Integration Across Images}
\label{fixationduration}
In our study, a stimulus is a collage with a set of images $I_i \in \mathbb{I}$.
During the search task, participants fixate on multiple images in the collage, which generates fixations $G_i \in \mathbb{G}$ for each image $I_i$.
Hence, we need a mechanism to aggregate information across images.
To do this, we propose a weighted average scheme of the computed posteriors per image: 
\begin{align}
p(C | \mathbb{I},\mathbb{G}) = \sum_i \frac{d_j}{\sum_j d_j} p(C | I_i, G_i).
\end{align}
We consider for the weights $d_i$ the total fixations duration on image $I_i$ as well as fixed $d_i$ (see~\autoref{fig:short}). The latter corresponds to plain averaging.
\subsection{Attended Class Activation Mapping}
In order to inspect the internal representation of our Gaze Pooling Layer, we propose the attended class activation map visualization. It highlights discriminative image regions for a hypothesized search target based on CNN features combined with the weights from the gaze data. In this vein, it shares similarities to the CAM of ~\cite{zhou2016cvpr} but incorporates the gaze information as attention scheme.
The key idea is to delay the average pooling, which allows us to show spatial maps as also illustrated in \autoref{fig:short}. In more detail, our network consists of several convolutional layers which the features of last convolutional layer is weighted by our fixation density map (GWFM). We do global average pooling over the GWFM and use those features for a fully connected layer to get the user attended categories or attributes. Given that our features maps are weighted by gaze data of users, it represents their attended classes. We can identify the importance of the image region for attended categories by projecting back the weights of the output layer onto a gaze-weighted convolutional feature map, which we call Attended Class Activation Map (ACAM): 
\begin{align}
ACAM_c(x,y)=\sum_k w_{k}^{c}~GWFM_{k}(I,G)
\end{align}
where $w_{k}^{c}$ indicates the importance $GWFM_{k}(I,G)$ of unit k for class c. The procedure for generating the class activation map are shown in \autoref{fig:short2}.
\subsection{Implementations Details }
\label{sec:imple}

In order to obtain the CNN features maps, we follow \cite{zhou2016cvpr} and build on the recent VGGnet-GAP model. 
For our categorization experiments, we fine-tune on a 10 class classification problem on the DeepFashion data set \cite{liuLQWTcvpr16DeepFashion}.
For attribute prediction, we fine-tune a model with 10 times 2-way classification in the final layer.
We perform a validation of the VGGnet image classification performance model in the same setting as \cite{liuLQWTcvpr16DeepFashion} and obtained comparable results ($\pm 5\% $) for category and attribute classification.
To ensure that the images and collages are not informative of the category or attribute search tasks, we have performed a sanity check by using only the CNN prediction on the images of our collages. The resulting performance is at chance level, which validates our setup as search task information cannot be derived from the images or collages and therefore can only come from the gaze data.

%% file: ExperimentsandResults.tex
\vspace{1cm}
\section{Experiments}

To evaluate our method for search target prediction of categories and attributes, we performed a series of experiments.
We first evaluated the effectiveness of our Gaze Pooling Layer, the effect of using a local vs global representation, and of using a weighting by fixation duration.
We then evaluated the gaze encoding that encompasses the pooling scheme of the individual fixation as well as the $\sigma_\text{fix}$ parameter to represent a single fixation.
Finally, we evaluated the robustness of our method to noise in the eye tracking data, which sheds light on different possible deployment scenarios and hardware that our approach is amendable to.
Additionally, we provide visualization of the internal representations in the Gaze Pooling Layer.
Across the results, we present Top-N accuracies denoting correct predictions if the correct answer is among the top N predictions.

\subsection{Evaluation of the Gaze Pooling Layer}
\label{sec:gaze_pooling}
Fixation information enters our method in two places: The fixation density maps in the Gaze Pooling Layer(~\autoref{gazepooling}) as well as the weighted average across images in the form of fixation duration~(see \autoref{fixationduration} and \autoref{fig:short}). 
In order to evaluate the effectiveness of our Gaze Pooling Layer, we evaluate two conditions: ``{\it local}'' makes full use of the gaze data and generates fixation density maps using the fixation location as described in our method section. ``{\it global}'' also generates a fixation density map, but does not use the fixation location information and therefore generates for each fixation a uniform weight across the whole fixated image.
In addition, we evaluate two more conditions, where we either used the fixation duration ({\scriptsize \Interval}) as a weight to the average class posterior of each fixated image (see \autoref{fixationduration}) or ignore the duration.

\renewcommand{\arraystretch}{1.3}
\begin{table}[!t]
\begin{center}
\begin{tabular}{@{}ccccccc@{}}\\
\toprule
Global vs. &  & \multicolumn{3}{c}{----------- Category --------} & Attribute\\ 
Local & {\scriptsize \Interval} & Top1 & Top2 & Top3 & Accuracy\\\hline
Global & & 31\%$\pm 5$ & 48\% $\pm 8$ & 62\% $\pm 8$& 20\%$\pm 1$\\
Local & & 49\%$\pm 7$ & 68\%$\pm 6$  & 78\%$\pm 6$ & 26 \%$\pm 1$\\
Global & \checkmark & 52\%$\pm 6$& 68\%$\pm 6$  & 78\%$\pm 6$ & 25\%$\pm1$\\
Local & \checkmark &{\bf 57\%$\pm$8} & {\bf 74\%$\pm$7} & {\bf 84\%$\pm$4} & {\bf 34\%$\pm$1}\\
\bottomrule
\end{tabular}
\end{center}
\caption{Evaluation of global vs. local gaze pooling with and without weighting based on the fixation duration {\scriptsize \Interval}.}
\label{tab:best}
\end{table}
\autoref{tab:best} shows the result of all 4 combinations of these conditions, with the first column denoting if local or global information was used and the second column {\scriptsize \Interval} whether fixation duration was used.
Absolute performance of our best model using local information and fixation duration were 57\%, 74\%, and 84\% on top1-3 accuracy respectively for the categorization task and 34\% accuracy for attributes.
The results show a consistent improvement (16 to 18 pp for categories, 6 pp for attributes) across all measures and tasks going from a global to a local representation (first to the second row).
Adding the weighting by fixation duration yields another consistent improvement for both local and global approach (another 6 to 5 pp for categories). Our best method improves overall by 22 to 26 pp on the categorization task and 14 pp on the attributes.
The global method without fixation duration (first row) is in a spirit similar to \cite{sattar15_cvpr} -- although the specific application differs.
All further experiments will consider our best model (last row) as the reference and justify the parameter choices (average pooling, $sigma_\text{fix}$) by varying each parameter one by one.
\subsection{Evaluation of the Gaze Encoding}
\label{sec:gaze_encoding}
We then evaluated the gaze encoding that takes individual fixations as input and produces a fixation density map.
We first evaluated the representation of a single fixation that depends on the parameter$\sigma_\text{fix}$, followed by the pooling scheme that combines multiple fixations into fixation density maps.
\paragraph{Effects of Fixation Representation Parameter $f_{\sigma}$.}
The parameter $\sigma_\text{fix}$ controls the spatial extend of a single fixation in the fixation density maps as described in \autoref{attention}.
We determined an appropriate setting of this parameter to be $\sigma_\text{fix}=1.6$ in a pilot study to roughly match the eye tracker accuracy and analyzed here the influence on the overall performance by varying this parameter in a sensible range (given eye tracker accuracy and coarseness of feature map) from 1 to 2 as shown in \autoref{tab:fixcode_sigma}.
As can be seen from the Table, our method is largely insensitive to the investigated range of reasonable choices of this parameter and our choice of 1.6 is on average a valid choice within that range.
\renewcommand{\arraystretch}{1.3}
\begin{table}[!t]
\begin{center}
\begin{tabular}{@{}ccccccc@{}}
\toprule
$\sigma_\text{fix}\rightarrow$ & $1$ & 1.2 & 1.4 & 1.6 & 1.8 & 2 \\\hline
Top1 & 55\% & 54\% & 56\% & 56\% & 57\% & 57\% \\
Top2 & 74\% & 74\% & 74\% & 74\% & 74\% & 75\% \\
Top3 & 83\% & 84\% & 84\% & 85\% & 85\% & 84\% \\
\bottomrule
\end{tabular}
\end{center}
\caption{Evaluation of different gaze encoding schemes using different per-fixation $\sigma_\text{fix}$.}\label{tab:fixcode_sigma}
\end{table}
\paragraph{Fixation Pooling Strategies.}
We evaluated two options for how to integrate single fixations into an fixation density map: Either using average or max pooling.
The results are shown in \autoref{tab:fixcode_pool}.
As the Table shows, while both options perform well, average pooling consistently improves over the max pooling option.

\renewcommand{\arraystretch}{1.3}
\begin{table}[!t]
\begin{center}
\begin{tabular}{@{}ccccc@{}}\\
\toprule
Fixation & \multicolumn{3}{c}{----------- Category --------} & Attribute\\ 
Pooling & Top1 & Top2 & Top3 & Accuracy\\\hline
Max & 54\%$\pm 8$ & 73\%$\pm 9$  & 83\%$\pm 6$&32\%$\pm1$\\
Average & \textbf{57\%$\pm$8} & \textbf{74\%$\pm$7} & \textbf{84\%$\pm$4} & \textbf{34\%$\pm$1}\\
\bottomrule
\end{tabular}
\end{center}
\caption{Evaluation of different fixation pooling strategies using average or max pooling.}
\label{tab:fixcode_pool}
\end{table}

\subsection{Noise Robustness Analysis}
\begin{figure}
\begin{center}
\includegraphics[width=\linewidth]{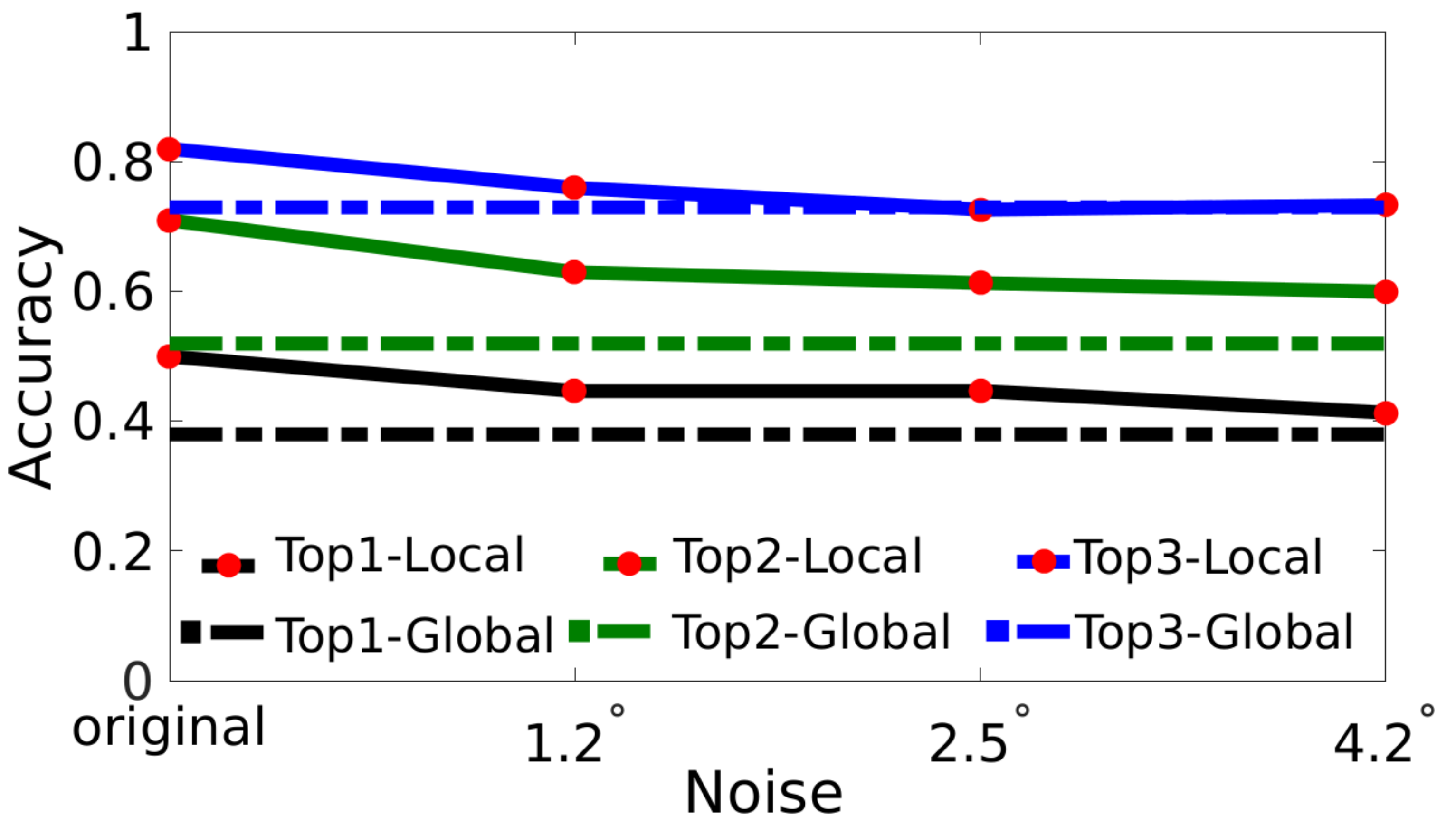}
\end{center}
\caption{Accuracy for different amounts of noise added to the eye tracking data. Our method is robust to this error which suggests that it can also be used with head-mounted eye trackers or learning-based methods that leverage RGB cameras integrated into phones, laptops, or public displays.}
\label{fig:noise}
\end{figure}

While our gaze data is recorded with a highly-accurate stationary eye tracker, there are different modalities and types of eye trackers available.
One key characteristic in which they differ is the error at which they can record gaze data -- typically measured in degrees of visual angle.
While our controlled setup provides us with an accuracy of about 0.7 degrees of error, state-of-the-art eye trackers based on webcams, tablets or integrated into glasses can have up to 4 degrees depending also on the deployment scenario~\cite{zhang15_cvpr}.
Therefore, we finally investigated the robustness of our approach w.r.t. different levels of (simulated) noise in the eye tracker.
To this end, we sampled noise from a normal distribution with $\sigma = 1, 3, 5$.
This corresponds roughly to $60, 120$ and $200$ pixels and to $1.2, 2.5$ and $4.2$ degrees of visual angles and hence covers a realistic range of errors.
The results of this evaluation are shown in \autoref{fig:noise}.
As can be seen, our method is quite robust to noise with only a drop of 5 to 10pp for Top3 to Top1 accuracy, respectively -- even at the highest noise level.
In particular, all the results are consistently above the performance of the corresponding global methods shown as dashed lines in the plot.
\begin{figure}
\begin{center}
\includegraphics[width=\linewidth]{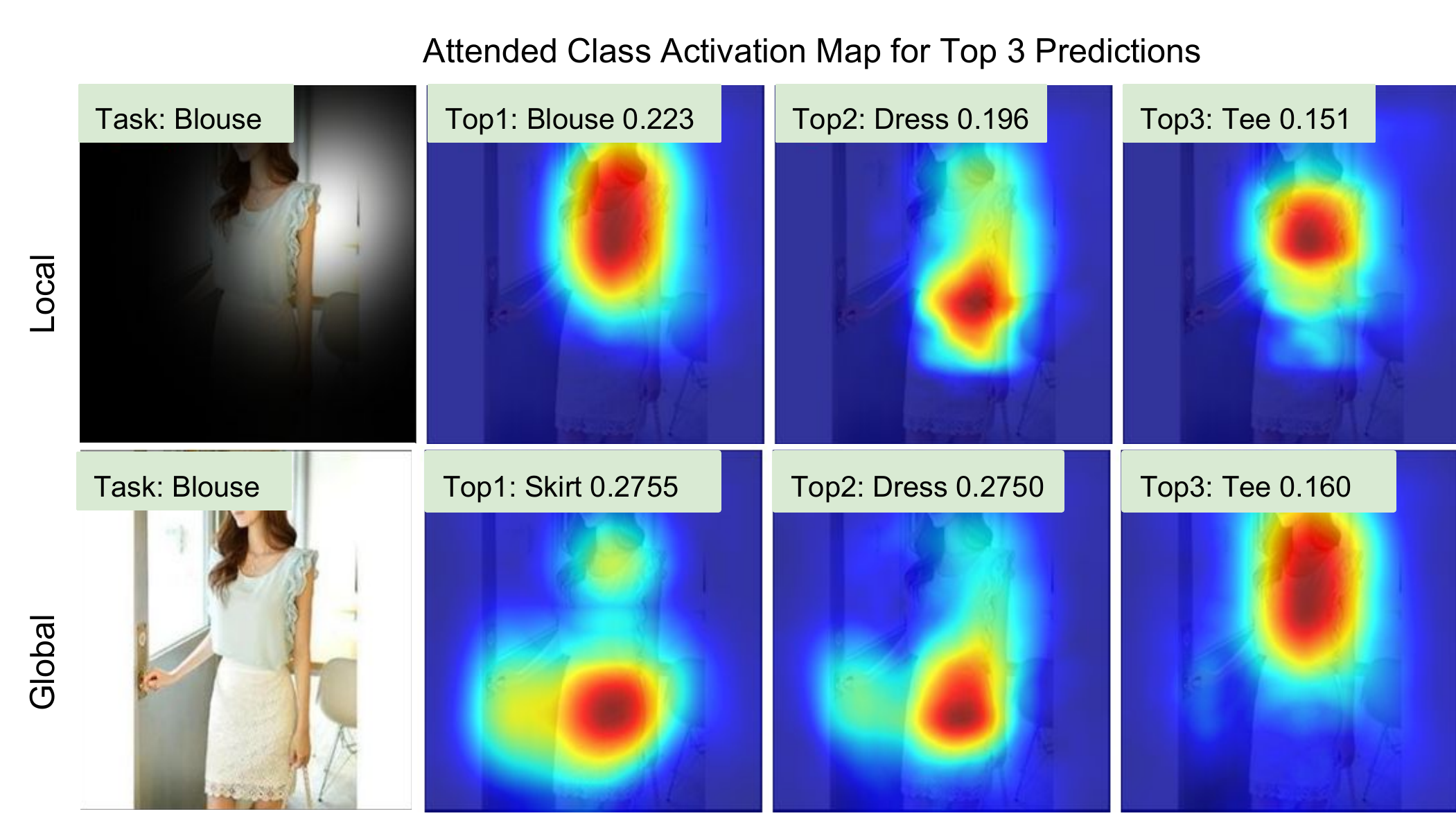}
\end{center}
\caption{Attended class activation maps of top 3 predictions in local and global method for a given image. Participants were searching for target category ``Blouse". The maps shows the discriminative image regions used for for this search task.}
\label{fig:visu}
\end{figure}

\subsection{Visualization and Analysis of Gaze Pooling Layer on Single Images}
We provide further insights into the working of our Gaze Pooling Layer by showing visual examples of the attended class activation maps, associated fixation density map and search target prediction results. While the quantitative evaluation was conducted on full collages, this is impracticable for inspection. Therefore, we show in the following visualizations and analysis on single images.  

\begin{figure*}[!t]
\begin{center}
\includegraphics[width=0.75\linewidth]{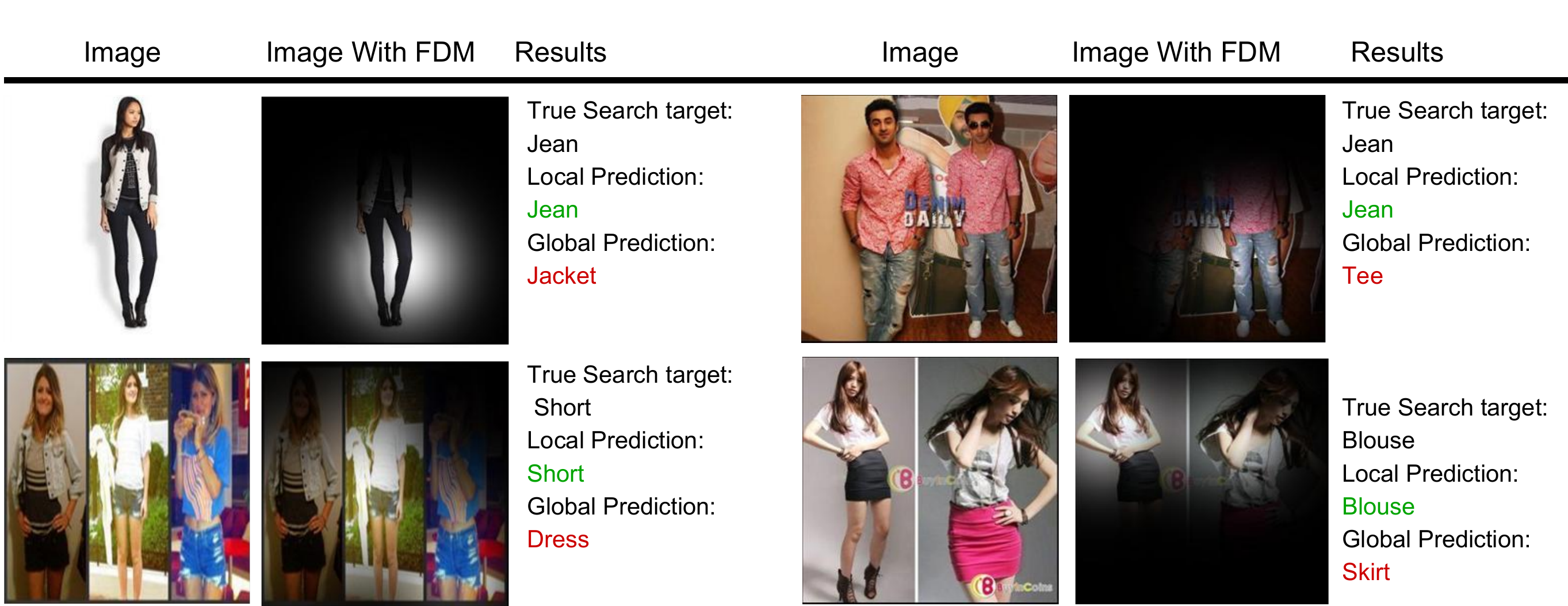}\\[0.1cm]
\end{center}
   \caption{Example category responses of the local and global method. Green means correct and red means wrong target prediction.}
\label{fig:vis}
\end{figure*}

\begin{figure*}[!t]
\begin{center}
\includegraphics[width=0.75\linewidth]{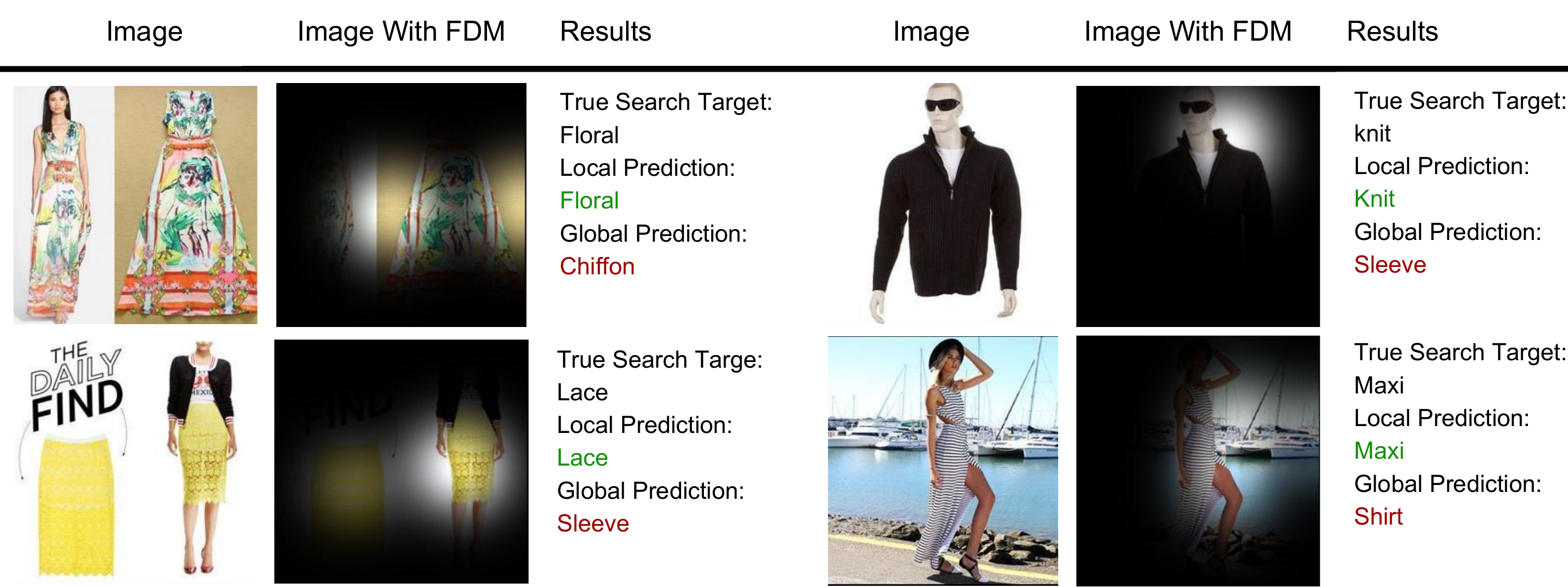}\
\end{center}
   \caption{Example attribute responses of local and global method. Green means correct and red means wrong target prediction.}
\label{fig:vis2}
\end{figure*}

\paragraph{Predictions.}
\autoref{fig:vis} shows results for the categorization task and \autoref{fig:vis2} for the attribute task.
Each of these figures shows the output of the ``global'' method that uses uniform fixation density map as well as the ``local'' method that makes full use of the gaze data. We observe that for the ``local'' method a relevant part of the images is fixated on which in turn leads to correct prediction of the intended search task.

\paragraph{Attended Class Activation Map (ACAM) Visualization.}
\autoref{fig:visu}~shows the attended class activation map (ACAM) of top 3 predictions, for ``local'' as well as ``global'' approach. The ``global'' method exploits that this image was fixed on - but does not exploit the location information of the fixations. Therefore it reduces in the case of a single image to a standard CAM. E.g the lower part of the image is activated for ``skirt'' and the upper part is activated for ``Tee''. One can see that highlighted regions vary across predicted class. The first row shows the ACAM for the ``local'' method. It can be seen how the local weighting due to the fixation is selective to the relevant features of the search target, e.g. eliminating the ``skirt'' responses and retaining the ``blouse'' responses.

\section{Discussion}

In this work, we studied the problem of predicting categories and attributes of search targets from gaze data.
\autoref{tab:best} shows strong performance for both tasks. 
Our Gaze Pooling Layer represents a modular and effective integration of visual and gaze features that are compatible with modern deep learning architectures.
Therefore, we would like to highlight three features that are of particular practical importance.
\paragraph{Parameter Free Integration Scheme.}
First, our proposed integration scheme is basically parameter-free. We introduce a single parameter $\sigma_\text{fix}$ but the gaze encoding is only input to the integration scheme and, in addition, the method turns out to be not sensitive to the choice (see experiments in \autoref{attention}).
\paragraph{Training from Visual Data.}
Second, fixing the fixation density maps to uniform maps yields a deep architecture similar to a GAP network that is well-suited for various classification tasks. 
While this no longer addresses the task of predicting categories and attributes intended by the human in the loop,
it allows us to train the remaining architecture for the task at hand and on visual data, which is typically easier to obtain in larger quantities than gaze data.
This type of training results in a domain-specific image encoding as well as a task-specific classifier.

\paragraph{Training Free Gaze Deployment.}
Gaze data is time-consuming to acquire -- which makes it rather incompatible with today's data hungry deep learning models.
In our model, however, the fixations density maps computed from the gaze data can be understood as spatially localized feature importance that is used to weight feature importance in the spatial image feature maps \autoref{fig:visu}.
Our results demonstrate that strong performance can be obtained with this re-weighting scheme without the need to re-train with gaze data.
As a result, our approach can be deployed without any gaze-specific training. This result is surprising, in particular as the visual model on its own is completely uninformative without gaze data on the task of search target prediction (as we have validated in \autoref{sec:gaze_pooling}. We believe this simplicity of deployment is a key feature that makes the use of gaze information in deep learning practical.

\begin{figure}[!t]
\centering
\includegraphics[width=0.8\columnwidth]{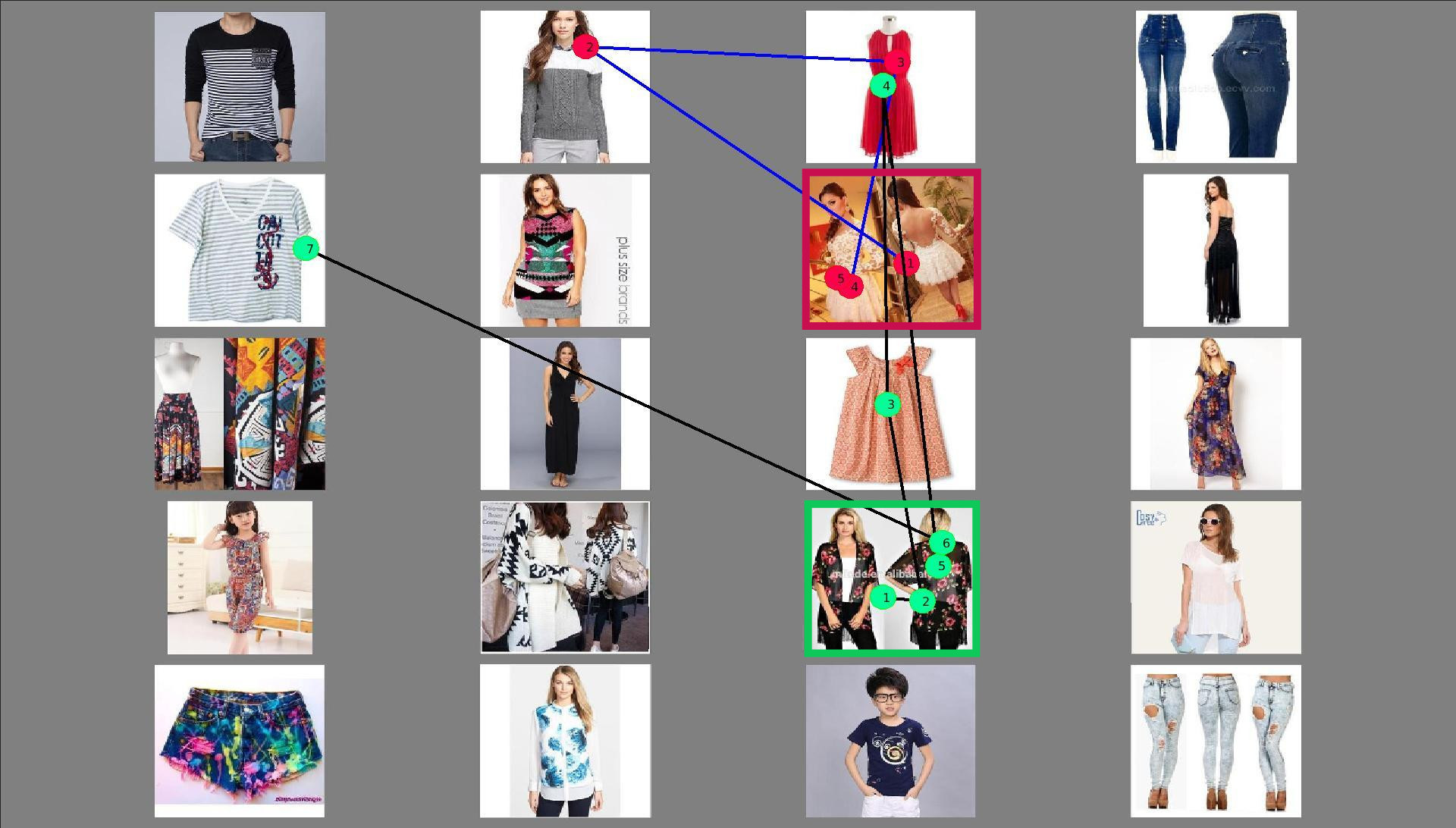}\\[0.1cm]
\includegraphics[width=0.8\columnwidth]{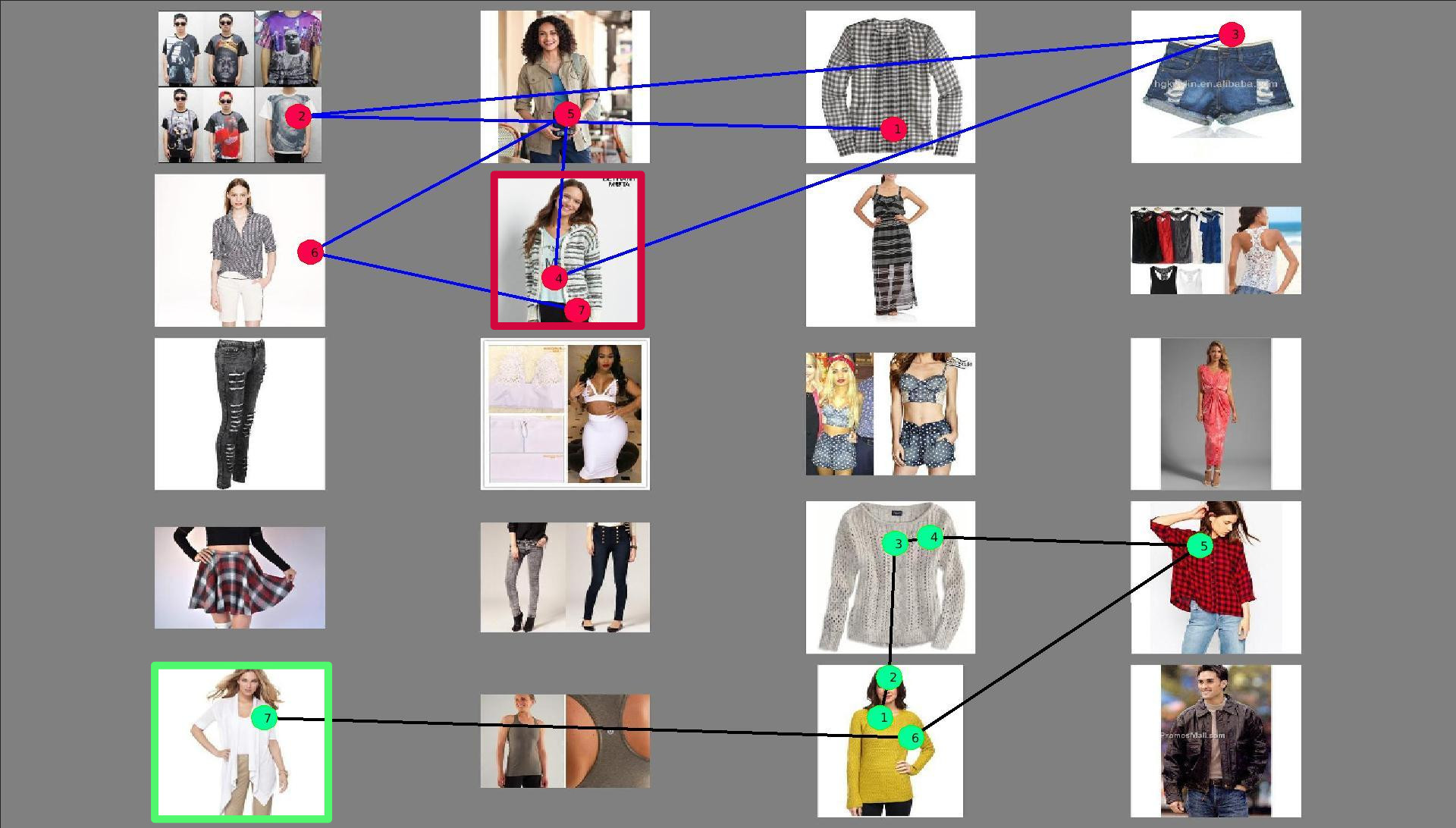}\\[0.1cm]
\caption{Example fixation data of 2 participants (red and green dots) with search target  attribute=`Floral' on top and category=`Cardigan' below.}
\label{fig:mentalpic}
\end{figure}

\paragraph{Biases in Mental Model of Attributes and Categories Among Users.}

In order to illustrate the challenges our Gaze Pooling Layer has to deal with in terms of the variations in the observed gaze data, we show example fixation data in \autoref{fig:mentalpic}. In each image, fixation data of two participants (red and green dots) is overlaid over a presented collage. Although both participants had the same search target (top: attribute `Floral'; bottom: category `Cardigan'), we observe a drastically different fixation behaviour. One possible explanation is that the mental models of the same target category or attribute can vary widely depending on personal biases \cite{4657362}. Despite these strong variations in the gaze information, our Gaze Pooling Layer allows predicting the correct answer in all 4 cases. The key to this success is aggregating relevant local visual feature across all images in the collage, that in turn represent one consistent search target in terms of categories and attributes.

%% file: Conclusions.tex
\section{Conclusion}

In this work, we proposed the first method to predict the category and attributes of visual search targets from human gaze data.
To this end, we proposed a novel Gaze Pooling Layer that allows us to seamlessly integrate semantic and localized fixation information into deep image representations.
Our model does not require gaze information at training time, which makes it practical and easy to deploy.
We believe that the ease of preparation and compatibility of our Gaze Pooling Layer with existing models will stimulate further research on gaze-supported computer vision, particularly methods using deep learning.

\subsection*{Acknowledgment}
This research was supported by the German Research Foundation (DFG CRC 1223) and the Cluster of Excellence on Multimodal Computing and Interaction (MMCI) at Saarland University. We also thank Mykhaylo Andriluka for helpful comments on the paper.

%% file: sup.tex
\section{Supplementary Material}
We present additional results that accompany the main paper submission. This results were omitted from the submission due to space constraints. 
In \autoref{collage} we show visualization of the fixations path on image collage, ACAM and predictions of our method for the images collages. 
\autoref{whitin}, represents within participants predictions. We show performance of our method over increasing number of fixations in \autoref{time}. 
Furthermore, we show more visualizations of ACAM for single images for different categories and attributes (\autoref{acam}). we show additional results to illustrate different scan path of users for shared search target. We also provide a \href{https://youtu.be/wPeyK5jkjUU}{{\bf video}}\footnote{Link to the video: https://youtu.be/wPeyK5jkjUU} of one search task and prediction over time for further understanding. 
\\
\\
\vspace{-1cm}
\subsection{Search target prediction over image collage}\label{collage}
\begin{figure*}[h]
\centering
\includegraphics[width=\linewidth]{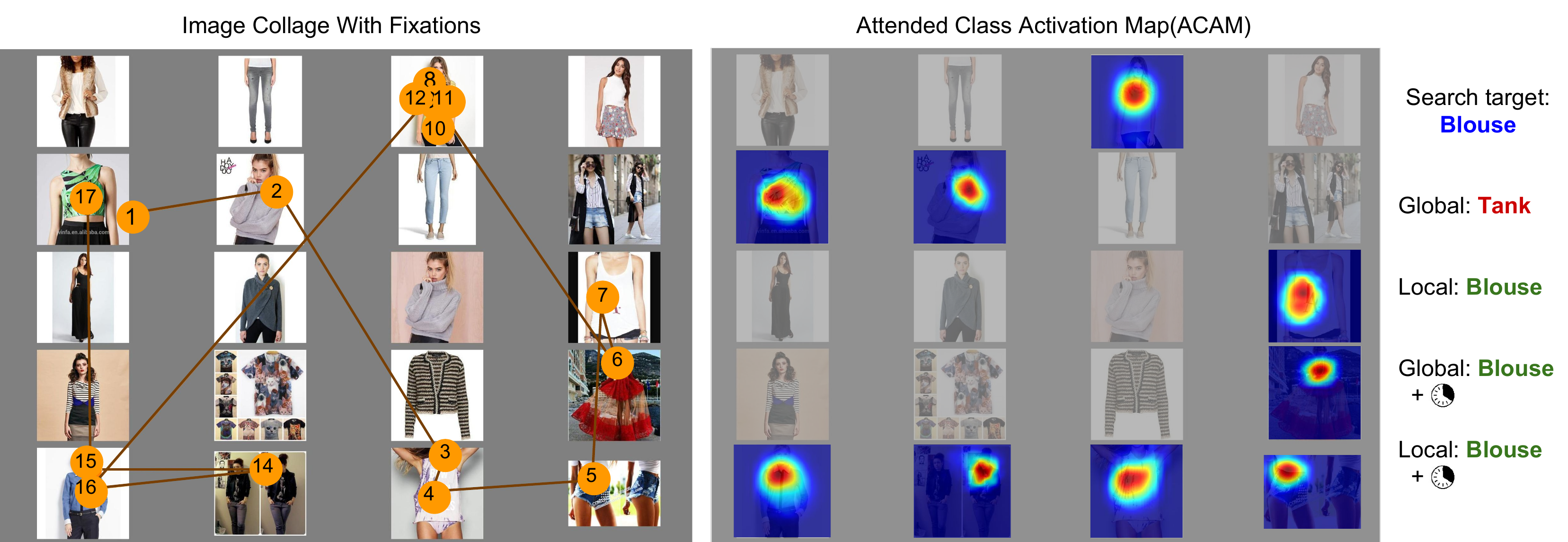}
\includegraphics[width=\linewidth]{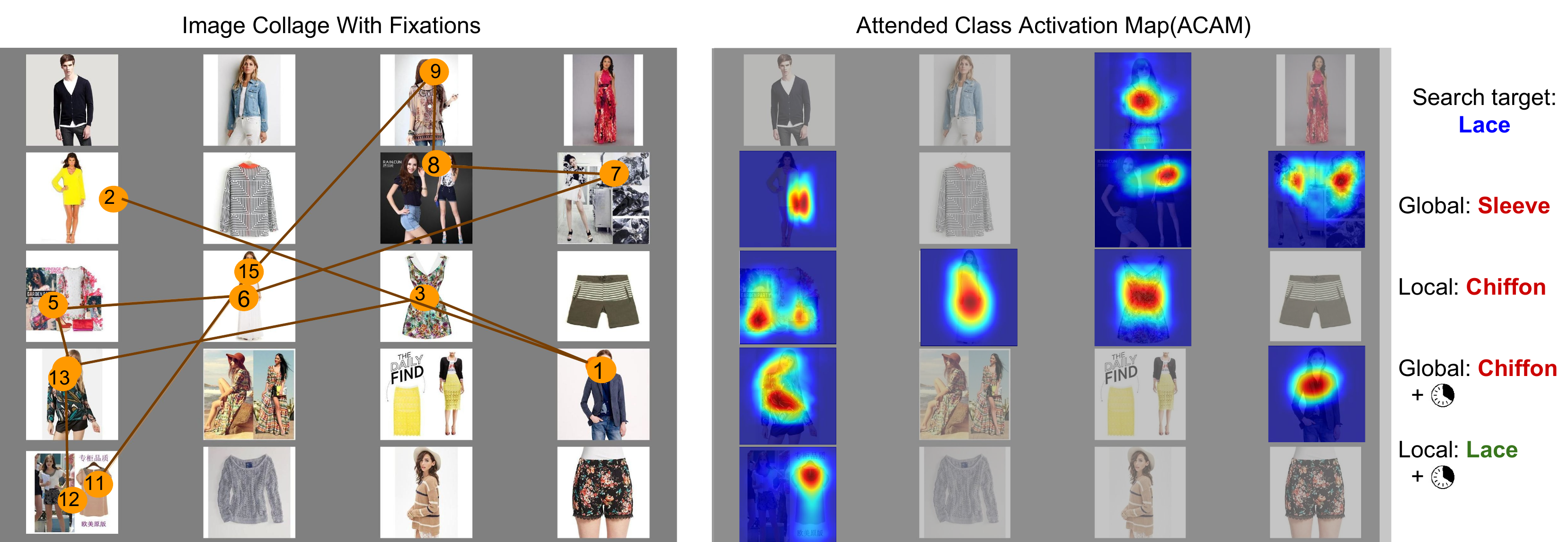}
\caption{Image collage with fixations of a participant searching for ``Blouse'' and ``Lace''. The right image show the ACAM of each fixated image in the collage. The last column represent top 1 prediction for global and local method without and with fixation durations.}
\label{fig:point2}
\end{figure*}

In \autoref{fig:point2}, we present the search target prediction over image collages for categories and attributes.
Top left image represent fixation data of one participant searching for category ``Blouse''. The bottom image represent fixation data of user searching for ``Lace''.The posterior of all fixated images are average to get one final prediction over all fixated images. This results are visualization of presented results in section 5.1 of the paper. 
\\
\\
\vspace{-1cm}
\subsection{Within User results}\label{whitin}
\begin{figure}[H]
\centering
\includegraphics[width=0.7\linewidth]{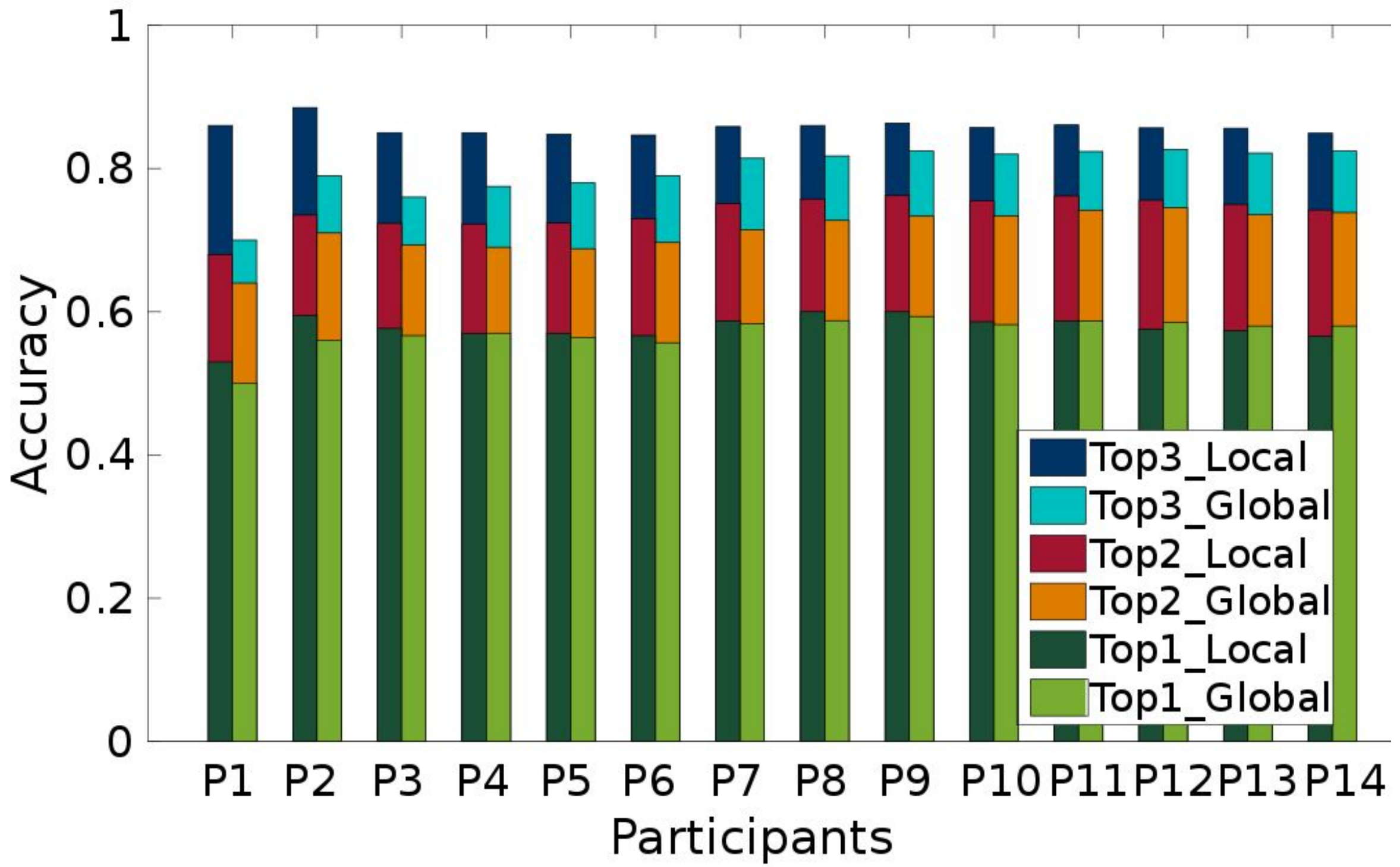}
\includegraphics[width=0.7\linewidth]{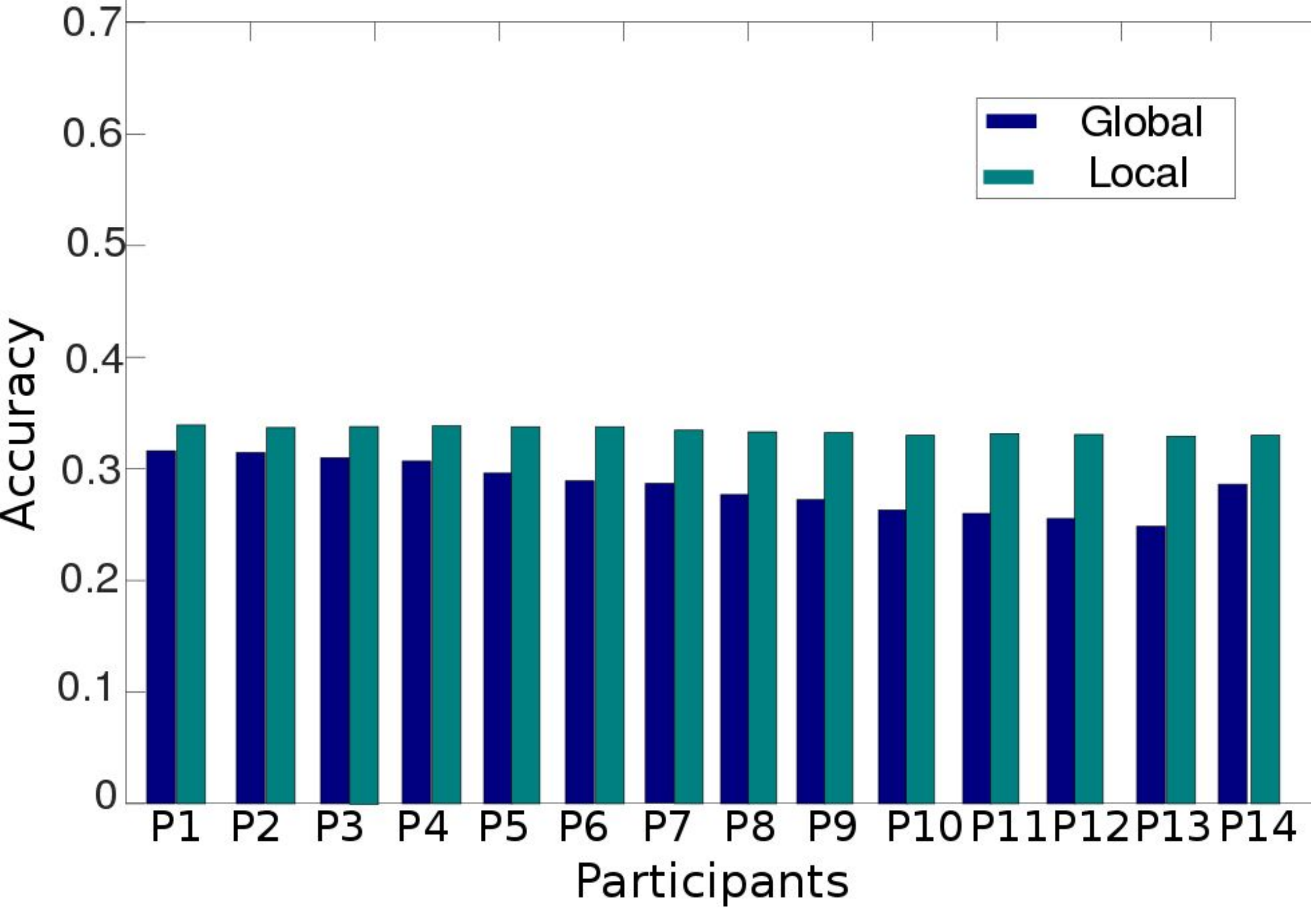}
\caption{With in user performance of our best local and global model for categories and attributes.The image on top shows top3 predictions of categories. Image on bottom represents the top1 attributes predictions.}
\label{fig:within}
\end{figure}

In the main paper we presents cross participants results (Section 5.1 main paper). \autoref{fig:within} shows results per participant. 
\subsection{Performance Over Time}\label{time}
\begin{figure}[H]
\centering
\includegraphics[width=0.9\linewidth]{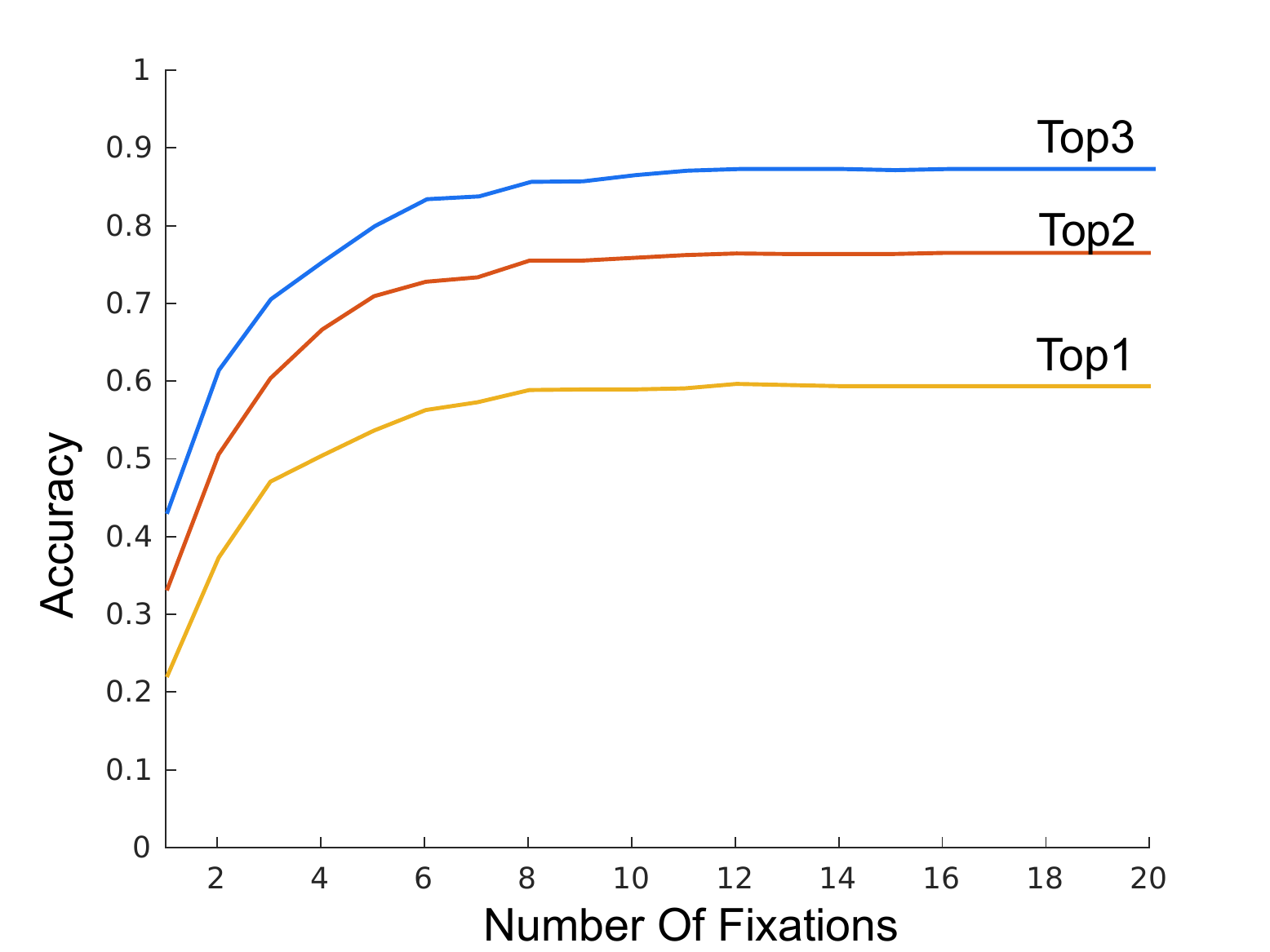}
\caption{Performance of our best model(local+fixation durations) over increasing number of fixations for categories search prediction}
\label{fig:time}
\end{figure}

\autoref{fig:time}, illustrate the effect of information accumulation over time. Participants performs various number of fixations on each image collage to find the search target. As in one image collage they only have 2 fixations in another 
they may perform over 10 fixations to find the right target. \autoref{fig:time} our goal was to show what total number of fixations could be useful to find the target. As one can see from figure after 8 fixations accuracies doesn't varies so much. 
\subsection{Attended Class Activation Map}\label{acam}
\autoref{fig:point1} shows the attended class activation map(ACAM) of top 1 predictions of each single fixated images, for ``local'' and ``global'' approach. 
The left most column represent the image and the task which is the search target, the second column shows fixation density maps of user searching for the given task and the last two columns are ACAM for local and global.
This results are additional visualizations, associated with section 5.4 in the main paper. 
\begin{figure*}
\includegraphics[width=\linewidth]{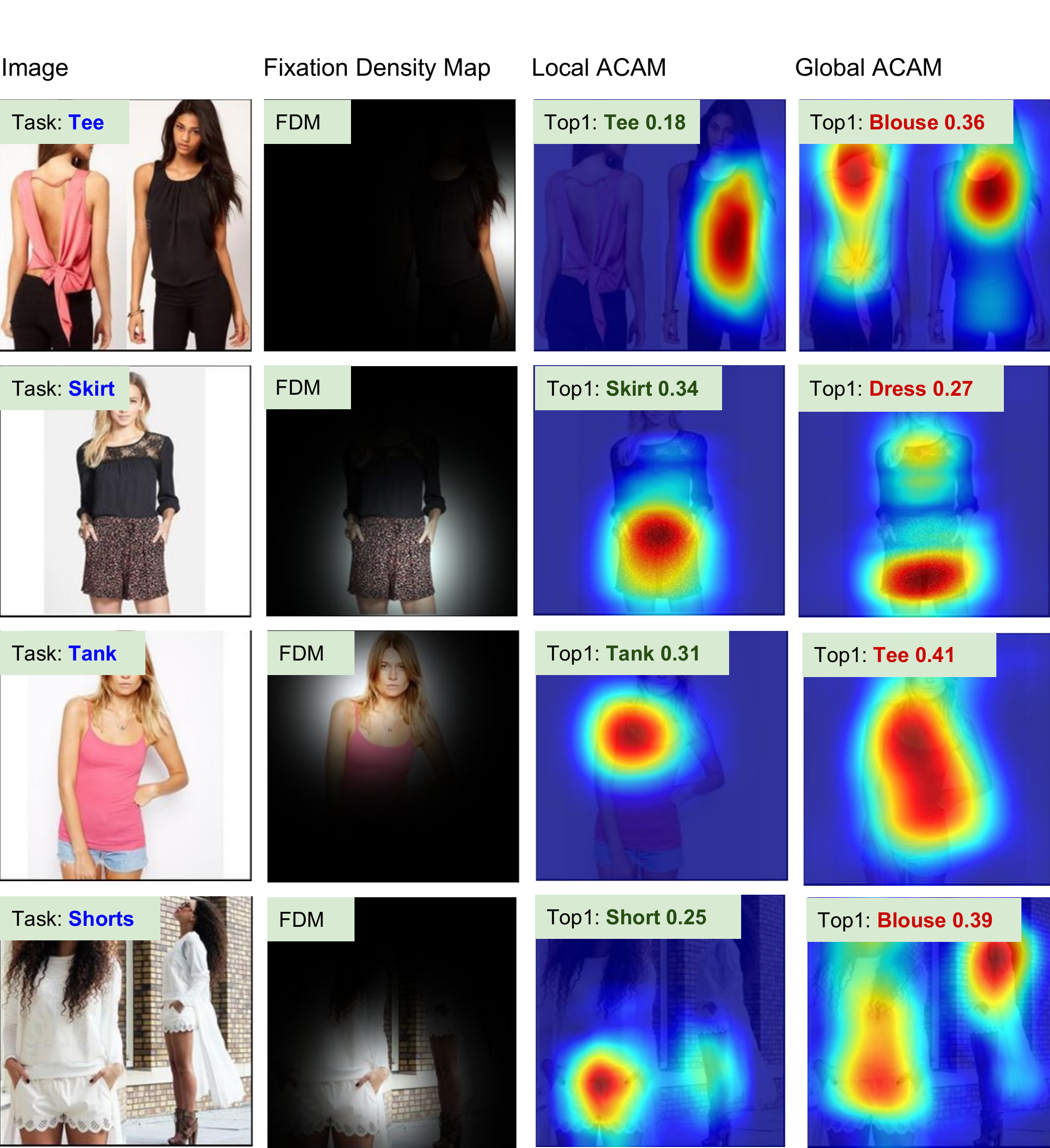}
\caption{Attended class activation maps of top1 prediction in local and global method for a single fixated image. 
Participants were searching for the given category. The maps shows the discriminative image regions used for this search task}
\label{fig:point1}
\end{figure*}
\autoref{fig:point1} shows the attended class activation map(ACAM) of top 1 predictions, for ``local'' and ``global'' approach. 
\begin{figure*}{h}
\includegraphics[width=\linewidth]{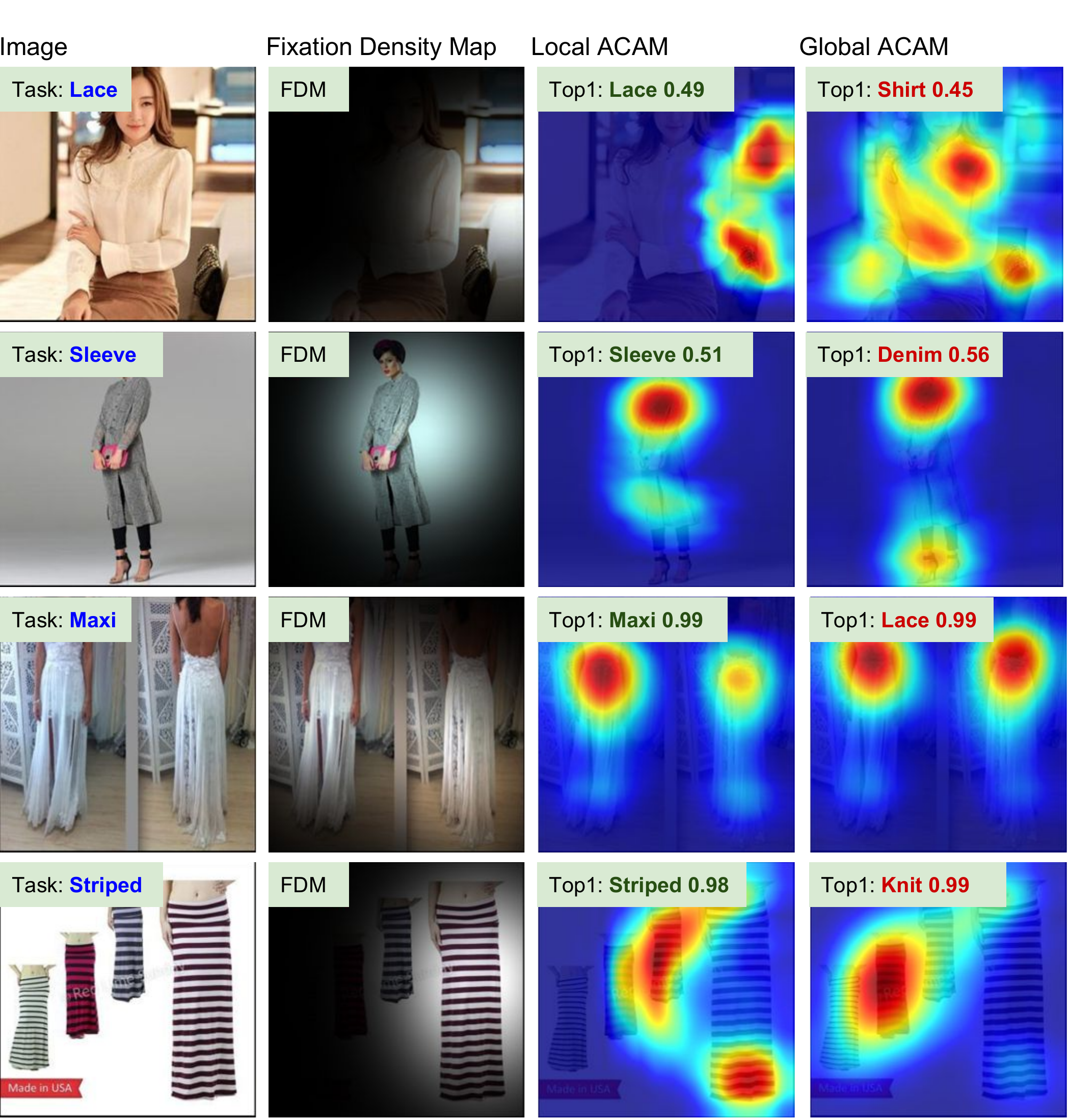}
\caption{Attended class activation maps of top1 prediction in local and global method for a single fixated image. Participants were searching for the given attribute. The maps shows the discriminative image regions used for this search task}
\label{fig:point1}
\end{figure*}
\subsection{Differences in fixations path for Attributes and Categories Among Users}\label{diff}
\begin{figure*}[h]
\centering
\includegraphics[angle=0, width=\linewidth]{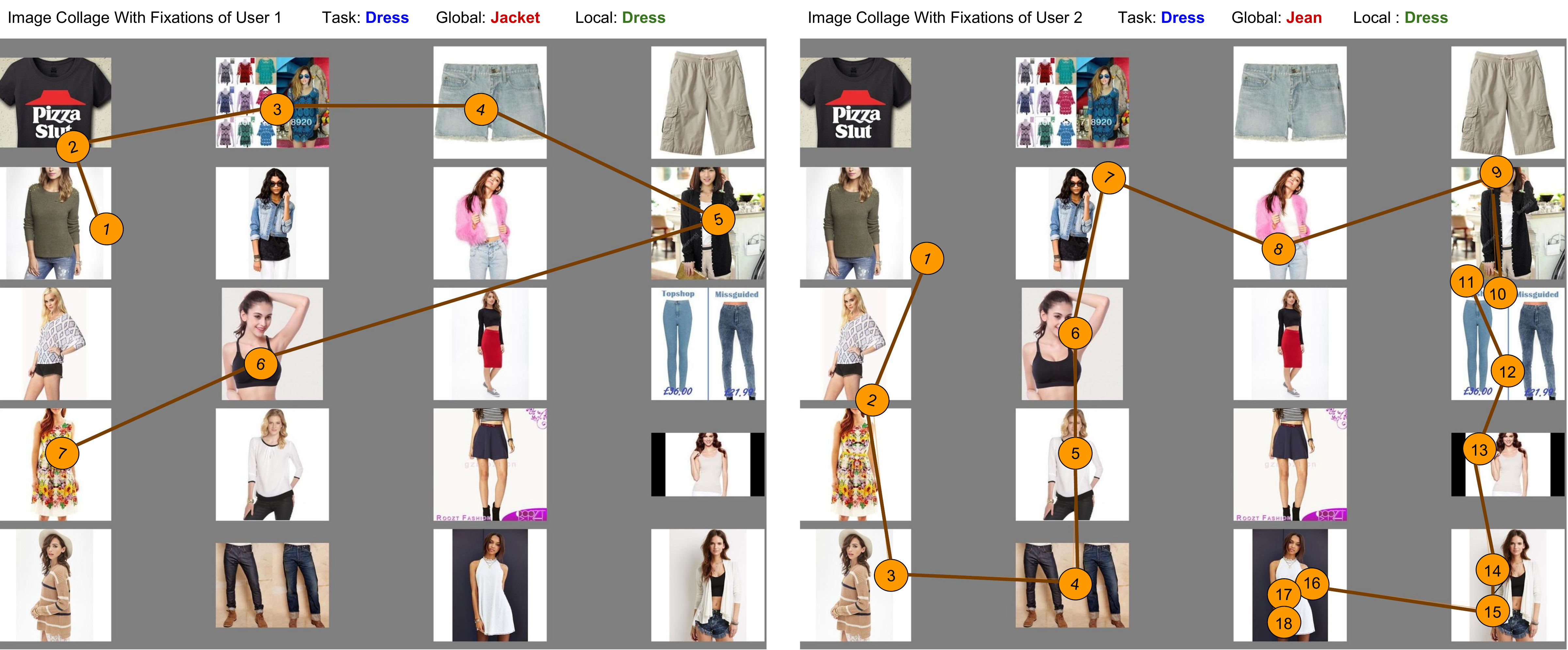}
\caption{Example fixation data of 2 participants (right and left image collage) with search target category ``Dress''.}
\label{fig:point5}
\end{figure*}

In \autoref{fig:point5}, shows example fixation data of two users searching for ``Dress''. Althought both participants had the same search target, their fixation behavior are different. Yet, our local approach manages to successfully predict the category of the search target. This figure are additional results, associated with section 6 in the main paper. Results, associated with section 6 in the main paper. 
